\documentclass{article}

\PassOptionsToPackage{numbers, compress}{natbib}
\usepackage[dblblindworkshop, final]{neurips_2025}
\workshoptitle{Evaluating the Evolving LLM Lifecycle}

\usepackage[utf8]{inputenc} 
\usepackage[T1]{fontenc}    
\usepackage{hyperref}       
\usepackage{url}            
\usepackage{booktabs}       
\usepackage{amsfonts}       
\usepackage{nicefrac}       
\usepackage{microtype}      
\usepackage{xcolor}         
\usepackage{graphicx} 
\usepackage{amsmath}
\usepackage{amsthm}
\usepackage{tabularx}
\usepackage{algorithm}
\usepackage{dsfont}
\usepackage{makecell}
\usepackage[noend]{algpseudocode}
\usepackage{enumitem}
\usepackage[T1]{fontenc}
\usepackage{tgcursor}
\usepackage{multirow}
\usepackage{svg}
\usepackage{minted}
\usepackage{listings}
\usepackage[most]{tcolorbox}
\tcbuselibrary{skins}
\usepackage{wrapfig}

\title{
No-Human in the Loop: Agentic Evaluation at Scale for Recommendation

}

\author{%
Tao Zhang\thanks{Equal contribution} \\
Walmart Global Tech \\
Sunnyvale, California, USA \\
\texttt{tao.zhang0@walmart.com} \\
\And
Kehui Yao\footnotemark[1] \\
Walmart Global Tech \\
Sunnyvale, California, USA \\
\texttt{kehui.yao@walmart.com} \\
\And
Luyi Ma\footnotemark[1] \\
Walmart Global Tech \\
Sunnyvale, California, USA \\
\texttt{luyi.ma@walmart.com} \\
\And
Jiao Chen\footnotemark[1] \\
Walmart Global Tech \\
Sunnyvale, California, USA \\
\texttt{jiao.chen0@walmart.com} \\
\And
Reza Yousefi Maragheh\footnotemark[1] \\
Walmart Global Tech \\
Sunnyvale, California, USA \\
\texttt{Reza.Yousefimaragheh@walmart.com} \\
\And
Kai Zhao\footnotemark[1] \\
Walmart Global Tech \\
Sunnyvale, California, USA \\
\texttt{kai.zhao@walmart.com}
\And
Jianpeng Xu \\
Walmart Global Tech \\
Sunnyvale, California, USA \\
\texttt{jianpeng.xu@walmart.com}
\And
Evren Korpeoglu \\
Walmart Global Tech \\
Sunnyvale, California, USA \\
\texttt{EKorpeoglu@walmart.com}
\And
Sushant Kumar \\
Walmart Global Tech \\
Sunnyvale, California, USA \\
\texttt{sushant.kumar@walmart.com}
\And
Kannan Achan \\
Walmart Global Tech \\
Sunnyvale, California, USA \\
\texttt{kannan.achan@walmart.com}
}

\tcbset{
  mybox/.style={
    enhanced,
    breakable,
    colback=gray!5,
    colframe=black!70,
    boxrule=0.5pt,
    sharp corners,
    fonttitle=\itshape,
    colbacktitle=white,   
    coltitle=black,
    width=\textwidth,
  },
  reportbox/.style={
    enhanced,
    breakable,
    colback=gray!5,
    colframe=black!70,
    boxrule=0.5pt,
    sharp corners,
    fonttitle=\itshape,
    colbacktitle=white,   
    coltitle=black,
    width=\textwidth,
  },
}
\newcommand{\system}{\textcolor{blue!70!black}{\textbf{system:}}}
\newcommand{\user}{\textcolor{blue!70!black}{\textbf{user:}}}

\begin{document}

\maketitle

\begin{abstract}
Evaluating large language models (LLMs) as judges is increasingly critical for building scalable and trustworthy evaluation pipelines. We present ScalingEval, a large-scale benchmarking study that systematically compares 36 LLMs—including GPT, Gemini, Claude, and Llama—across multiple product categories using a consensus-driven evaluation protocol. Our multi-agent framework aggregates pattern audits and issue codes into ground-truth labels via scalable majority voting, enabling reproducible comparison of LLM evaluators without human annotation. Applied to large-scale complementary-item recommendation, the benchmark uncovers several key takeaways: (i) Anthropic Claude-3.5-sonnet achieves the highest decision confidence, (ii) Gemini-1.5-pro offers the best overall performance across categories, (iii) GPT-4o provides the most favorable latency–accuracy-cost trade-off, and (iv) GPT-OSS-20B leads among open-source models. Category-level analysis further reveals strong consensus in structured domains (Electronics, Sports) but persistent disagreement in lifestyle categories (Clothing, Food). Together, these findings establish ScalingEval as a reproducible benchmark and evaluation protocol for LLMs-as-judges, offering both methodological advances and actionable insights into scaling, reliability, and model family trade-offs.


\end{abstract}

\section{Introduction}
\label{sec:introduction}

Large language models (LLMs) are rapidly emerging as evaluation engines across domains ranging from dialogue systems to recommender models. As their use expands, a central question arises: \textbf{can LLMs reliably serve as judges at scale, and how do different families compare in accuracy, efficiency, and robustness?} This question is especially pressing for high-stakes applications such as e-commerce recommendation, where evaluation quality directly impacts user trust and revenue.  

Complementary-Item Recommendation (CIR) provides a representative testbed for this challenge~\cite{hao2020p,
ma2021neat}. These models surface add-on items---e.g., suggesting a phone case with a smartphone---that must be complementary rather than substitutes or irrelevant products. Ensuring correctness requires capturing subtle linguistic and contextual cues, such as ``family-size vs. single-serve'' or ``battery-powered vs. corded.'' Traditional heuristics based on category overlap or co-purchase statistics remain computationally efficient, but they suffer from \textbf{contextual blind spots} and \textbf{static taxonomies} that cannot adapt to evolving product trends~\cite{sugahara2024really}. As a result, CIR evaluations often produce false positives and false negatives, degrading recommendation quality.  

Recent advances in LLMs promise a path forward. Their contextual reasoning and linguistic sensitivity are well suited to detecting nuanced product relationships. However, two obstacles remain: (i) na\"ively applying a single LLM across millions of item pairs is prohibitively expensive, and (ii) little is known about how judgments vary across model families, sizes, and settings. Without systematic benchmarks, it is unclear which models are most trustworthy, which categories are most challenging, or how agreement scales with model diversity.  
\vspace{-8pt}
\begin{wrapfigure}{r}{0.65\textwidth}
    \centering
    \includegraphics[width=0.63\textwidth]{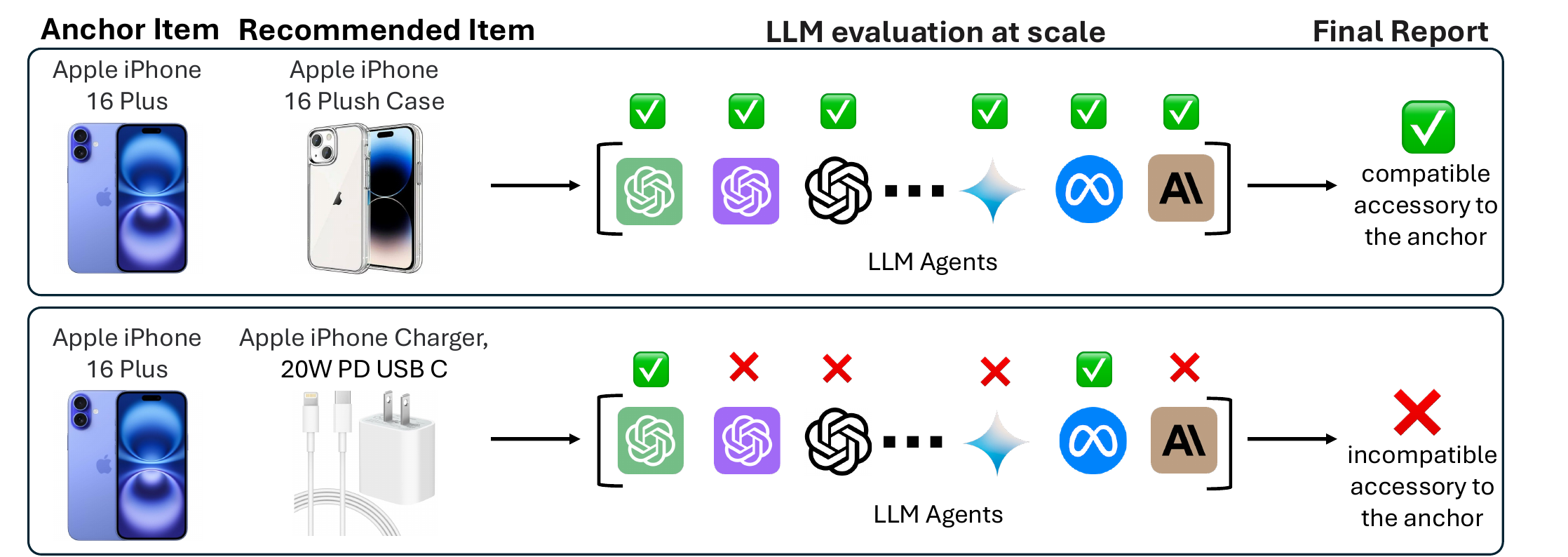}
    \vspace{-8pt}
    \caption{LLM-Agentic Evaluation at Scale without human in the loop}
    \label{fig:placeholder}
\end{wrapfigure}
\vspace{-8pt}

To address these gaps, we introduce \textbf{ScalingEval}, a large-scale, multi-agent benchmarking framework that positions LLMs themselves as judges. ScalingEval decomposes the evaluation task into specialized audit agents, integrates conflict resolution with strict prioritization rules, and synthesizes consensus ground truth via majority voting across 36 models spanning closed-source (GPT-4o, Claude-3.5-sonnet, Gemini-1.5/2.5) and open-source families (GPT-OSS-20B, Llama-3 variants). Applied to Walmart-scale CIR data across multiple product categories, ScalingEval reveals several key findings: \textbf{(i) Claude-3.5-sonnet delivers the highest decision-making confidence ($\sim$99\%), (ii) Gemini-1.5-pro achieves the best overall accuracy, coverage and latency, (iii) GPT-4o offers the most favorable latency--accuracy--cost trade-off, and (iv) GPT-OSS-20B emerges as the strongest open-source model.} Moreover, structured domains such as \textit{Electronics} and \textit{Sports} exhibit strong cross-model agreement, while lifestyle categories such as \textit{Food} and \textit{Clothing} remain challenging, underscoring the need for domain-sensitive evaluation protocols.  

By combining agentic orchestration with systematic benchmarking, ScalingEval contributes both a \textbf{reproducible methodology for LLM-as-a-judge evaluation} and \textbf{empirical insights into scaling behaviors, trade-offs, and agreement dynamics}. Beyond CIR, the framework provides a template for lifecycle evaluation of LLMs, supporting future research on emergent abilities, domain robustness, and cross-model reliability.

\section{Methodology}\label{sec:method}

\begin{figure}[htbp]
  \centering  
    \includegraphics[width=1.0\linewidth]{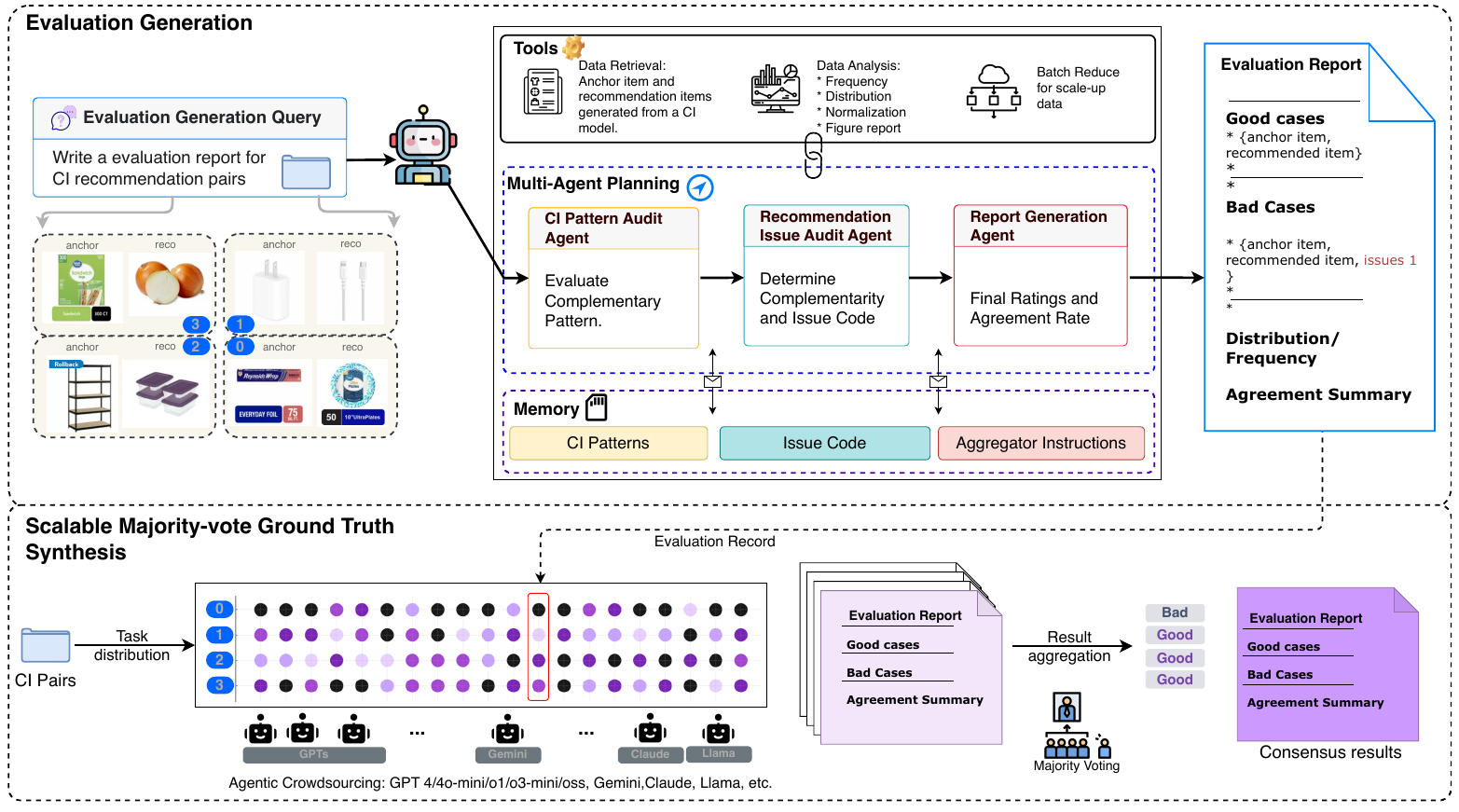} 
    \vspace{-15pt}
    \caption{The overview of ScalingEval framework.}
    \label{fig: Framework}
\end{figure}

In this section, we propose ScalingEval in Figure \ref{fig: Framework}, an LLM-based agentic framework for evaluating CIR. The framework orchestrates multiple specialized agents to conduct structured audits, detect common evaluation issues, and resolve disagreements through consensus mechanisms. By automating both ground-truth generation and multi-model comparison, ScalingEval enables large-scale, low-cost, and reproducible assessment without relying on human annotators.


\subsection{Agentic Evaluation Report Generation}
\label{subsec: Agentic Evaluation Report Generation}
A user query (e.g., “generate a CI report”) triggers a multi-agent pipeline that audits item pairs, resolves conflicts, and aggregates results, ensuring scalable and interpretable evaluation over datasets $\mathbf{D}={(item_i^a,, item_i^r)}_{i=1}^N$.
Both acceptance and rejection rubrics are defined to capture valid and invalid relations.

\noindent \textbf{CI Pattern Audit.} A state-of-the-art LLM proposes CI patterns $\mathbf{P}={p_i}_{i=1}^M$, validated via empirical use cases. Each pair $(item^a, item^r)$ is mapped to one or more patterns:
\[
out_{\text{pattern\_audit}} = f_{\text{pattern\_audit}}(item^a, item^r) = \{(item^a, item^r): [p_j]\}, \quad [p_j] \in \mathbf{P}.
\]

\noindent \textbf{Recommendation Issue Audit.} Using predefined issue codes $\mathbf{S}={s_i}_{i=1}^K$, the agent identifies reasons pairs may fail:
\[
out_{\text{issue\_audit}} = f_{\text{issue\_audit}}(item^a, item^r) = \{(item^a, item^r): [s_j]\}, \quad [s_j] \in \mathbf{S},
\]

\noindent \textbf{Report Generation.} Passed pairs are split into chunks $C_i$ of size $k$, 
\[
D = \bigcup_{i=1}^{M} C_i, \quad 
C_i = \{x_{(i-1)k+1}, \dots, x_{ik}\}, \quad 
M = \left\lceil \tfrac{N}{k} \right\rceil, \quad 
r_i = f_{\text{LLM}}(C_i)
\]
where each $r_i$ is a summary containing counts, pattern breakdowns, flagged issues, and conflicts:  
$
r_i = \{t_i, c_i, n_i, P_i, I_i, q_i, a_i\}.
$ 

Here,  
$t_i$: total pairs in chunk; 
$c_i$: complementary count;
$n_i$: non-complementary count; 
$P_i$: CI pattern; 
$I_i$: flagged issue codes; 
$q_i$: conflicted pairs;
$a_i$: agreement score.
Aggregation yields
\[
R = \{
T = \sum_{i=1}^M t_i,
C = \sum_{i=1}^M c_i, 
N = \sum_{i=1}^M n_i,
Q = \sum_{i=1}^M q_i, 
P = \bigoplus_{i=1}^M P_i, 
I = \bigoplus_{i=1}^M I_i, 
A = \bigoplus_{i=1}^M a_i
\}
\]   
with the constraint:  
$
C + N = T.
$

\subsection{Scalable Majority-vote Ground Truth Synthesis}
\label{subsec: Consensus-Driven Judgement Generation}
As shown in the bottom of Figure~\ref{fig: Framework}, we strengthen report reliability through multi-agent verification and majority-vote synthesis. Multiple LLMs (e.g., GPT, Gemini, LLaMA) independently run the evaluation process in Section~\ref{subsec: Agentic Evaluation Report Generation}, producing judgments for each anchor–recommendation pair.

\definecolor{Reject}{RGB}{0,0,0}        
\definecolor{Major}{RGB}{229,207,251}    
\definecolor{Minor}{RGB}{197,163,250}    
\definecolor{Good}{RGB}{129,57,179}    

\noindent \textbf{Majority-Vote Ground Truth Synthesis}
collects individual judgments into a structured matrix where each row corresponds to an anchor–recommendation pair and each column corresponds to an agent’s decision.
Conflicting outputs are resolved using majority voting. 
If disagreement persists, a conflict-resolution policy 
(e.g., 
(\textcolor{Reject}{Reject} $>$ 
\textcolor{Major}{Major} $>$ 
\textcolor{Minor}{Minor} $>$ 
\textcolor{Good}{Good}) 
ensures conservative decisions.

\noindent \textbf{Judgement Aggregation} The consensus-based labels are further aggregated to measure agreement levels across agents.
Disagreements are logged as conflicted pairs, providing transparency into model uncertainty and dataset ambiguity.
The voting step ensures that systematic biases from any single LLM are reduced, and stability is achieved through collective reasoning.
Among multiple candidate reports generated, the Consensus Report is selected for completeness and consistency across the dataset.

\section{Experimental Results}
\label{sec: Experiments}


This section identifies which LLM offers the best balance between accuracy, coverage, and efficiency, while additional details on our Majority-Voting agreements, case study, experimental setup and demo prompt design are provided in the Appendix.

Table~\ref{tab: main results} summarizes model family performance, reporting average accuracy and coverage across three temperature settings.
First, let's answer the best performer in accuracy vs. latency trade-offs in closed-source group:
(1) High accuracy, large latency: gpt-o1 model;
(2) Balanced performance: gemini-1.5-pro, gpt-4o;
(3) Short latency, lower accuracy: gemini-1.5-flash, claude-3.5-sonnet.
Gemini-1.5-pro is the top performer overall, gpt-4o also offers balanced speed and accuracy but cost less, and GPT-OSS leads open-source models with mid-tier closed-source performance at lower cost.
For category-specific performance, gemini-1.5-pro leads in 4 out of 5 categories. 
GPT-o1 excels in \textit{Clothes \& Shoes}, while gpt-4o provides balanced performance with minimal latency.  
By domain,\textit{Sports \& Outdoors} and \textit{Food \& Beverages} reach the highest accuracy and coverage. \textit{Electronics} and \textit{Home \& Garden} show steady results across top models. \textit{Clothes \& Shoes} stands out with unique patterns, suggesting room for further gains.

\vspace{-5pt}
\begin{table*}[htbp]
\centering
\renewcommand\arraystretch{1.3}
\resizebox{\textwidth}{!}{%
\begin{tabular}{lcccccccccccccc}
\toprule
\multirow{2}{*}{\textbf{LLM}} & 
\multicolumn{2}{c}{\textbf{Overall}} & 
\multicolumn{2}{c}{\textbf{Electronics}} &
\multicolumn{2}{c}{\textbf{Home \& Garden}} &  
\multicolumn{2}{c}{\textbf{Sport \& Outdoors}} & 
\multicolumn{2}{c}{\textbf{Closes \& Shoes}} & 
\multicolumn{2}{c}{\textbf{Food \& Beverages}} & 
\multirow{2}{*}{\textbf{Latency}} & 
\multirow{2}{*}{\textbf{Cost/pair}}\\
\cmidrule(lr){2-3} \cmidrule(lr){4-5} \cmidrule(lr){6-7} \cmidrule(lr){8-9}
\cmidrule(lr){10-11} \cmidrule(lr){12-13}
 & Accuracy & Coverage
 & Accuracy & Coverage
 & Accuracy & Coverage
 & Accuracy & Coverage
 & Accuracy & Coverage
 & Accuracy & Coverage
 & \\
\midrule
\multicolumn{14}{c}{\textbf{Closed Source}} \\
\midrule
gpt-4o  & 60.13   & 64.63   & 58.91  & 63.63  & 54.44  & 57.41  & \underline{60.89}  & \underline{62.09}  & \underline{78.15}  & \underline{79.83}  & 65.30  & 74.63  & 1x & 1x\\
gpt-4o-mini & 34.49   & 37.04   & 36.36  & 38.91  & 30.37  & 31.85  & 18.12  & 19.76  & 23.53  & 28.57  & 54.85  & 58.58  & 1.5x & 1x\\
gpt-o1  & \underline{67.18}   & \underline{68.81}   & \underline{77.09}  & \underline{81.82}  & \underline{67.41}  & \underline{70.37}  & 20.97  & 22.58  & \textbf{82.35}  & \textbf{89.91}  & \underline{74.63}  & \underline{79.85}  & 4.9x & 6.0x\\
gpt-o3-mini & 39.40   & 41.67  & 39.27  & 40.00  & 31.11  & 33.33  & 26.62  & 27.42  & 53.78  & 58.82  & 50.00  & 52.61  & 3.4x & 1.2x\\

gemini-2.5-pro  & 57.15   & 62.64   & 57.45  & 62.18  & 57.41  & 68.81  & 58.46  & 62.50  & 62.23  & 72.27  & 58.21  & 73.51  & 6.6x & 3x\\
gemini-2.0-flash  & 19.55   & 22.25  & 20.91  & 27.64  & 17.04  & 18.89  & 14.52  & 15.32  & 12.61  & 18.49  & 22.76  & 28.93  &  0.5x & 0.03x\\
gemini-1.5-pro  & \textbf{76.63}  & \textbf{87.14}   & \textbf{80.36}  & \textbf{85.45}  & \textbf{69.26}  & \textbf{81.85} & \textbf{87.50}   & \textbf{91.94}  & 63.87  & 68.91  & \textbf{80.60}  & \textbf{95.90}  &  1.1x & 1.5x\\
gemini-1.5-flash  & 34.56   & 40.58   & 36.36  & 41.45  & 33.33  & 39.26  & 18.95  & 26.61  & 20.01  & 21.85  & 53.73  & 61.94  &  0.4x & 0.03x\\

claude-3.5-sonnet  & 31.08   &31.58   & 33.45  & 34.19  & 25.56  & 25.93  & 19.35  & 19.35  & 31.10  & 32.78  & 40.67  & 40.67  & 0.9x & 1.5x\\
\midrule
\multicolumn{14}{c}{\textbf{Open Source}} \\
\midrule
gpt-oss-20b  & 42.32   & 52.60   & 38.55  & 46.18  & 42.96  & 52.96  & 39.11  & 47.58  & 38.66   & 50.42  & 48.89  & 56.34  & 5.9x$_{o}$  & 0.0x\\
llama3-3b-instruct  & 14.32   & 15.43   & 17.82   & 20.00   & 12.96  & 13.70   &7.26   &8.06   & 11.76  & 12.60   & 19.03  & 20.52  & 1.7x$_{o}$  & 0.0x\\
llama3-8b-instruct  & 17.58   & 18.14   & 19.64  & 20.72  & 13.70  & 14.07  & 10.89  & 11.29  & 13.45  & 13.45  & 26.12  & 26.12  & 1x$_{o}$ & 0.0x\\
\midrule
\bottomrule
\end{tabular}%
}
\vspace{-8pt}
\caption{Overall performance on all LLMs across multiple product categories.}
\label{tab: main results}
\end{table*}
\vspace{-5pt}

Table~\ref{tab:model_performance} reports confidence on pairs with definitive judgments (“Good” or “Bad”), excluding conflicts and unknowns. Results form clear tiers: Claude-3.5-sonnet leads (98.4–99.2\%), followed by Llama-3-8B-Instruct (95.9–97.5\%). GPT-4o (93.1–95.1\%) balances cost and confidence, while o1 models (94.3–95.5\%) deliver strong confidence with top coverage. Gemini models (87.9–88.6\%) trade confidence for breadth, and Llama-3-3B-Instruct achieves efficient but lower coverage (86.3–93.3\%). Several families (Gemini, Claude-3.5-sonnet, o1, o3) show minimal sensitivity to temperature.

\vspace{-5pt}
\begin{table*}[htbp]
\centering
\renewcommand\arraystretch{1.0}
\scriptsize
\resizebox{\textwidth}{!}{%
\begin{tabular}{lrrrrrrl}
\toprule
\textbf{Model} & \textbf{Confidence} & \textbf{Kappa} & \textbf{Issue Match} & \textbf{Coverage} & \textbf{F1} & \textbf{Agreement} & \textbf{Top Categories Performance} \\
\midrule
anthropic\_claude-3.5-sonnet\_temp\_0.8 & 99.20 & 98.40 & 99.20 & 31.20 & 99.20 & 99.20 & Electronics: 98.20, Sports: 97.80, Pets: 97.50 \\
anthropic\_claude-3.5-sonnet\_temp\_0.6 & 98.80 & 97.60 & 98.80 & 31.60 & 98.80 & 98.80 & Electronics: 98.00, Sports: 97.50, Pets: 97.20 \\
anthropic\_claude-3.5-sonnet\_temp\_0.4 & 98.40 & 96.70 & 98.40 & 31.60 & 98.40 & 98.40 & Electronics: 97.80, Sports: 97.20, Pets: 96.80 \\
meta\_Llama\_3\_8B\_Instruct\_temp\_0.6 & 97.50 & 95.00 & 97.50 & 18.10 & 97.50 & 97.50 & Electronics: 96.50, Sports: 95.80, Pets: 95.20 \\
meta\_Llama\_3\_8B\_Instruct\_temp\_0.4 & 97.40 & 94.90 & 97.40 & 17.60 & 97.40 & 97.40 & Electronics: 96.20, Sports: 95.50, Pets: 94.80 \\
meta\_Llama\_3\_8B\_Instruct\_temp\_0.8 & 95.90 & 91.80 & 95.90 & 18.70 & 95.90 & 95.90 & Electronics: 95.80, Sports: 94.20, Pets: 93.50 \\
openai\_o1\_temp\_0.6 & 95.50 & 91.00 & 95.50 & 68.80 & 95.50 & 95.50 & Electronics: 94.50, Sports: 93.80, Pets: 93.20 \\
openai\_gpt-4o-mini\_temp\_0.4 & 95.10 & 90.10 & 95.10 & 36.50 & 95.10 & 95.10 & Electronics: 94.20, Sports: 93.50, Pets: 92.80 \\
openai\_o1\_temp\_0.8 & 94.40 & 88.80 & 94.40 & 71.40 & 94.40 & 94.40 & Electronics: 93.80, Sports: 93.20, Pets: 92.50 \\
openai\_o1\_temp\_0.4 & 94.30 & 88.70 & 94.30 & 72.50 & 94.30 & 94.30 & Electronics: 93.50, Sports: 92.80, Pets: 92.20 \\
openai\_gpt-4o-mini\_temp\_0.6 & 93.90 & 87.80 & 93.90 & 37.00 & 93.90 & 93.90 & Electronics: 93.20, Sports: 92.50, Pets: 91.80 \\
openai\_gpt-4o\_temp\_0.4 & 93.90 & 87.80 & 93.90 & 64.10 & 93.90 & 93.90 & Electronics: 93.00, Sports: 92.20, Pets: 91.50 \\
openai\_gpt-4o-mini\_temp\_0.8 & 93.80 & 87.60 & 93.80 & 36.30 & 93.80 & 93.80 & Electronics: 92.80, Sports: 92.00, Pets: 91.20 \\
openai\_o3-mini\_temp\_0.6 & 93.70 & 87.30 & 93.70 & 41.70 & 93.70 & 93.70 & Electronics: 92.50, Sports: 91.80, Pets: 91.00 \\
openai\_gpt-4o\_temp\_0.6 & 93.50 & 87.10 & 93.50 & 64.60 & 93.50 & 93.50 & Electronics: 92.20, Sports: 91.50, Pets: 90.80 \\
openai\_o3-mini\_temp\_0.8 & 93.50 & 87.00 & 93.50 & 41.60 & 93.50 & 93.50 & Electronics: 92.00, Sports: 91.20, Pets: 90.50 \\
openai\_o3-mini\_temp\_0.4 & 93.40 & 86.90 & 93.40 & 43.10 & 93.40 & 93.40 & Electronics: 91.80, Sports: 91.00, Pets: 90.20 \\
llama\_3\_2\_3B\_Instruct\_temp\_0.6 & 93.30 & 86.70 & 93.30 & 15.40 & 93.30 & 93.30 & Electronics: 91.50, Sports: 90.80, Pets: 90.00 \\
openai\_gpt-4o\_temp\_0.8 & 93.10 & 86.20 & 93.10 & 64.20 & 93.10 & 93.10 & Electronics: 91.20, Sports: 90.50, Pets: 89.80 \\
llama\_3\_2\_3B\_Instruct\_temp\_0.4 & 92.80 & 85.60 & 92.80 & 13.40 & 92.80 & 92.80 & Electronics: 90.80, Sports: 90.20, Pets: 89.50 \\
gemini\_2\_0\_flash\_temp\_0.8 & 92.20 & 84.30 & 92.20 & 21.40 & 92.20 & 92.20 & Electronics: 90.50, Sports: 89.80, Pets: 89.20 \\
gemini\_2\_0\_flash\_temp\_0.6 & 89.60 & 79.20 & 89.60 & 22.30 & 89.60 & 89.60 & Electronics: 89.20, Sports: 88.50, Pets: 87.80 \\
gemini\_2\_0\_flash\_temp\_0.4 & 88.60 & 77.20 & 88.60 & 21.50 & 88.60 & 88.60 & Electronics: 88.80, Sports: 88.20, Pets: 87.50 \\
gemini\_gemini-2.5-pro\_temp\_0.6 & 88.60 & 77.20 & 88.60 & 62.60 & 88.60 & 88.60 & Electronics: 88.50, Sports: 87.80, Pets: 87.20 \\
gemini\_gemini-2.5-pro\_temp\_0.8 & 88.30 & 76.60 & 88.30 & 66.60 & 88.30 & 88.30 & Electronics: 88.20, Sports: 87.50, Pets: 86.80 \\
gemini\_1\_5\_pro\_temp\_0.6 & 88.10 & 76.20 & 88.10 & 87.10 & 88.10 & 88.10 & Electronics: 87.80, Sports: 87.20, Pets: 86.50 \\
gemini\_gemini-2.5-pro\_temp\_0.4 & 88.10 & 76.20 & 88.10 & 64.90 & 88.10 & 88.10 & Electronics: 87.50, Sports: 86.80, Pets: 86.20 \\
gemini\_1\_5\_pro\_temp\_0.8 & 88.10 & 76.20 & 88.10 & 86.40 & 88.10 & 88.10 & Electronics: 87.20, Sports: 86.50, Pets: 85.80 \\
gemini\_1\_5\_pro\_temp\_0.4 & 87.90 & 75.80 & 87.90 & 87.70 & 87.90 & 87.90 & Electronics: 86.80, Sports: 86.20, Pets: 85.50 \\
llama\_3\_2\_3B\_Instruct\_temp\_0.8 & 86.30 & 72.50 & 86.30 & 18.70 & 86.30 & 86.30 & Electronics: 85.50, Sports: 84.80, Pets: 84.20 \\
gemini\_1\_5\_flash\_temp\_0.6 & 85.70 & 71.50 & 85.70 & 40.60 & 85.70 & 85.70 & Electronics: 85.20, Sports: 84.50, Pets: 83.80 \\
gemini\_1\_5\_flash\_temp\_0.8 & 85.50 & 71.00 & 85.50 & 40.40 & 85.50 & 85.50 & Electronics: 84.80, Sports: 84.20, Pets: 83.50 \\
gemini\_1\_5\_flash\_temp\_0.4 & 85.40 & 70.90 & 85.40 & 40.20 & 85.40 & 85.40 & Electronics: 84.50, Sports: 83.80, Pets: 83.20 \\
gpt-oss-20b\_temp\_0.6 & 81.80 & 63.60 & 81.80 & 52.60 & 81.80 & 81.80 & Electronics: 82.50, Sports: 81.80, Pets: 81.20 \\
gpt-oss-20b\_temp\_0.8 & 78.50 & 57.00 & 78.50 & 52.90 & 78.50 & 78.50 & Electronics: 79.20, Sports: 78.50, Pets: 77.80 \\
gpt-oss-20b\_temp\_0.4 & 77.70 & 55.40 & 77.70 & 54.20 & 77.70 & 77.70 & Electronics: 78.80, Sports: 78.20, Pets: 77.50 \\
\bottomrule
\end{tabular}
} 
\vspace{-8pt}
\caption{Model performance by Confidence (desc.) across metrics and categories }
\label{tab:model_performance}
\end{table*}
\vspace{-5pt}

\section{Conclusion}
\label{sec: conclusion}
ScalingEval shows that large language models can act as reliable, no-human-in-the-loop judges for complementary-item recommendation by combining pattern audits, issue detection, and multi-agent consensus. Our results reveal clear performance tiers—highlighting Gemini-1.5-Pro as most accurate and efficient, Claude-3.5-Sonnet as most confident, and GPT-OSS-20B as the strongest open-source option—offering a path toward scalable, trustworthy evaluation of recommender systems.

\clearpage
\bibliographystyle{unsrtnat}
\bibliography{ref}

\clearpage
\appendix

\section{Experimental Setup}
\label{subsec:Experimental Setup}
\noindent\textbf{Datasets}. We consider real-world customer behaviors from Walmart e-commerce platform and sample data from seven major product categories: (1) \textit{Electronics}, (2) \textit{Sports \& Outdoors}, (3) \textit{Pet Supplies}, (4) \textit{Home \& Garden}, (5) \textit{Toys \& Games}, (6) \textit{Food \& Beverages}, and (7) \textit{Clothing \& Shoes}. We analyze anchor-recommendation pairs where each pair consists of a base product and a recommended complementary item, with a total of 1,745 pairs across all categories. The data includes evaluation results from 36 different AI models across multiple temperature settings (0.4, 0.6, 0.8), covering major model families such as Anthropic Claude, OpenAI GPT, Meta Llama, and Google Gemini. We also report the Agreement Rate to indicate the model consensus complexity for each category: the higher agreement rate means models show more consistent evaluation patterns and clearer recommendation criteria in that dataset.

\noindent\textbf{Implementation Details}. We select a diverse set of 36 state-of-the-art language models as the backbone models. The evaluation framework processes each anchor-recommendation pair through all available models, generating appropriateness judgments ("Good"/"Bad"/"Conflict"/"Unknown") based on pattern recognition and audit results. For all LLM-related methods, each experiment is evaluated on the complete dataset, with results aggregated across temperature variations and model families. We follow the majority voting approach for ground truth creation, requiring 60\% agreement threshold from the top 25 models based on quality scores. The system uses precision metrics to evaluate model performance, calculating both overall precision and coverage across different product categories. All the experiments with open-source models are conducted with NVIDIA A100-SXM4-80GB$\times$2.

\noindent\textbf{Evaluation Metrics}. For each anchor-recommendation pair and the associated model evaluations, we calculate agreement rates across all 36 models to construct the consensus labels, and the ground truth is derived from majority voting among high-quality models. The evaluation framework assesses model performance using three key precision metrics: (1) \textbf{Accuracy} (accuracy across all pairs, counting undetermined/covered pairs as incorrect), 
(2) \textbf{Confidence} (accuracy on pairs where models made a determination), 
(3) \textbf{Coverage} (percentage of pairs that models made a determination),
(4) \textbf{Kappa}: cohen's Kappa coefficient measuring agreement between model predictions and ground truth, 
(5) \textbf{Issue Match}: how well the model identifies specific issues in $\textbf{S}$ with recommendations, and 
(6) \textbf{Agreement} = (Number of models agreeing with majority) / (Total number of valid models with determination "Good" or "Bad"). The system identifies high-agreement categories like \textit{Sports \& Outdoors} (93.8\%) and \textit{Electronics} (91.3\%) versus challenging categories like \textit{Clothing \& Shoes} (84.2\%) and \textit{Food \& Beverages} (85.4\%), providing insights into where recommendation systems need improvement.

\noindent \textbf{Base LLMs}. Including 9 closed-source LLMs, that is, 
GPT-4o/4o-mini/o1/o3-mini (\cite{achiam2023gpt, hurst2024gpt, jaech2024openai}), 
Claude-3.5-sonnet (\cite{details-about-metr-s-preliminary-evaluation-of-claude-3-5-sonnet}), 
Gemini-2.5-pro/2.0-flash/1.5-pro/1.5-flash (\cite{comanici2025gemini, team2024gemini, team2023gemini}), 
as well as 3 open-source LLMs, that is, 
gpt-oss-20b\footnote{https://huggingface.co/openai/gpt-oss-120b}, 
llama3-3b-instruct \footnote{https://huggingface.co/meta-llama/Llama-3.2-3B-Instruct}
and llama3-8b-instruct \footnote{https://huggingface.co/meta-llama/Meta-Llama-3-8B-Instruct} \cite{dubey2024llama}.
\section{Majority Voting Agreements}
\label{subsect:Why Majority Voting Trustworthy}
\begin{figure}[htbp]
  \centering  
    \includegraphics[width=1.0\linewidth]{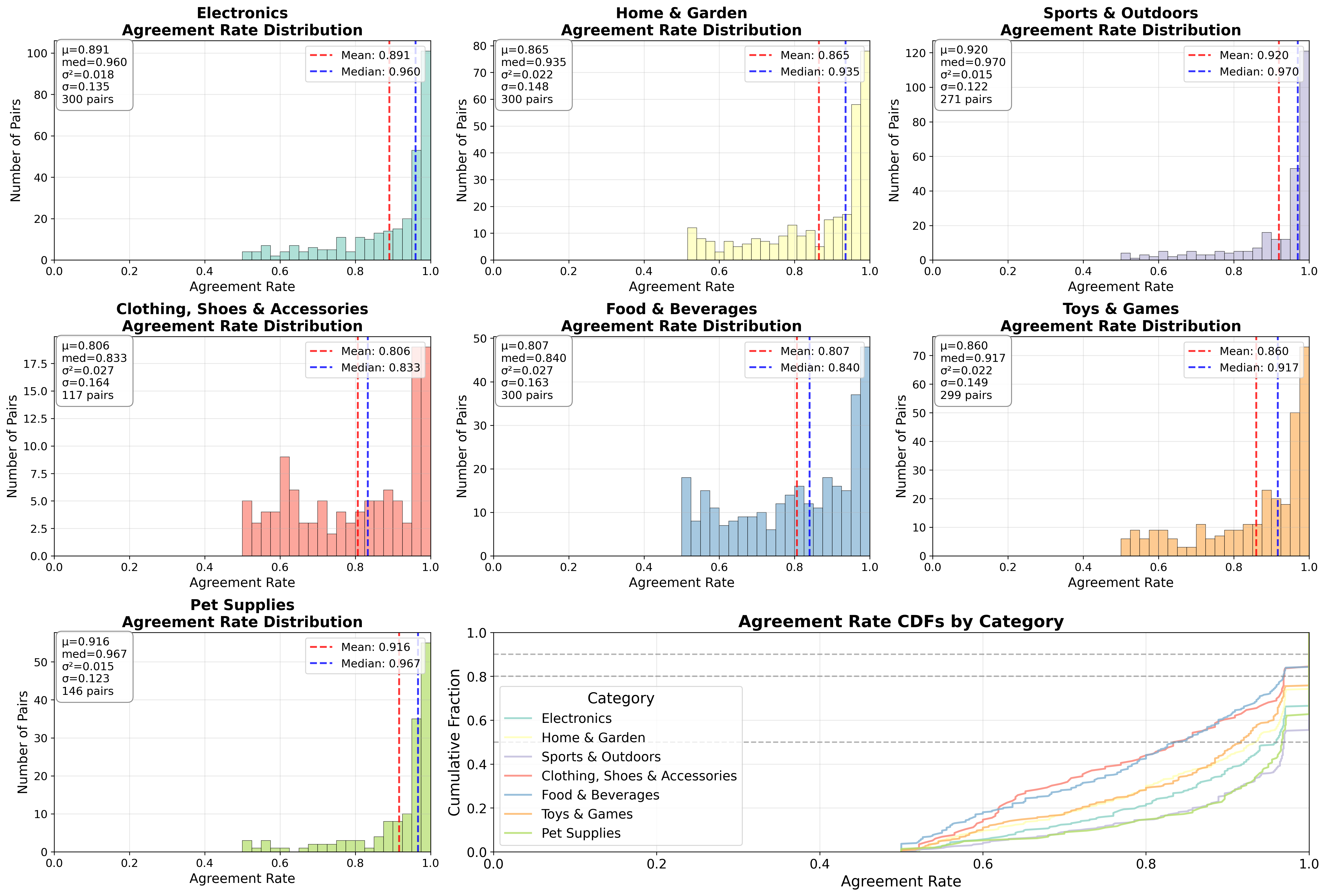} \caption{Different category agreement rate under pair frequency and cumulative fraction.}
    \label{fig: aggreement rate}
\end{figure}

This section examines the effectiveness of majority-voting ground truth synthesis as a consensus-based labeling approach that simulates human annotation.
The agreement rate distribution varies across categories, as shown in Figure~\ref{fig: aggreement rate}, highlighting differences in decision-making complexity under majority voting.

The histogram plot provides a group view of agreement rate distributions across all product categories.
Bar height shows the number of pairs at each agreement rate, while the distribution shape indicates variability: left-skewed = lower agreement, right-skewed = higher agreement, and bell-shaped = normal distribution around the mean.
The distribution analysis highlights distinct performance patterns across categories. 
\textit{Sports \& Outdoors} and \textit{Pet Supplies} show right-skewed distributions with most pairs above 0.8, reflecting high agreement and low variability. 
\textit{Electronics} also demonstrates good agreement, concentrated above 0.7, with moderate variability. 
\textit{Home \& Garden} and \textit{Toys \& Games} display moderate right-skewed or balanced distributions, indicating consistent but less concentrated agreement. 
In contrast, \textit{Clothing \& Shoes} and \textit{Food \& Beverages} exhibit more spread-out distributions, reflecting moderate agreement and higher variability.

The combined CDF (Cumulative Distribution Function) plot provides a single, comprehensive view of agreement rate distributions across all product categories. 
This visualization enables direct comparison of model consensus patterns and performance across different product domains. 
Steeper and right-shifted curves indicate stronger agreement, while higher curves reflect better overall performance. 
In contrast, left-shifted curves suggest lower agreement rates across pairs.
Across categories, \textit{Electronics}, \textit{Sports \& Outdoors}, and \textit{Pet Supplies} exhibit steep, high curves, indicating strong agreement and top performance. Lifestyle categories, including \textit{Clothing \& Shoes} and \textit{Food \& Beverages}, show flatter, lower curves, reflecting weaker performance. \textit{Home \& Garden} and \textit{Toys \& Games}, display intermediate patterns, balancing consistency with moderate accuracy.

\section{Case Study}

\begin{figure}[htbp]
  \centering  
    \includegraphics[width=0.8\linewidth]{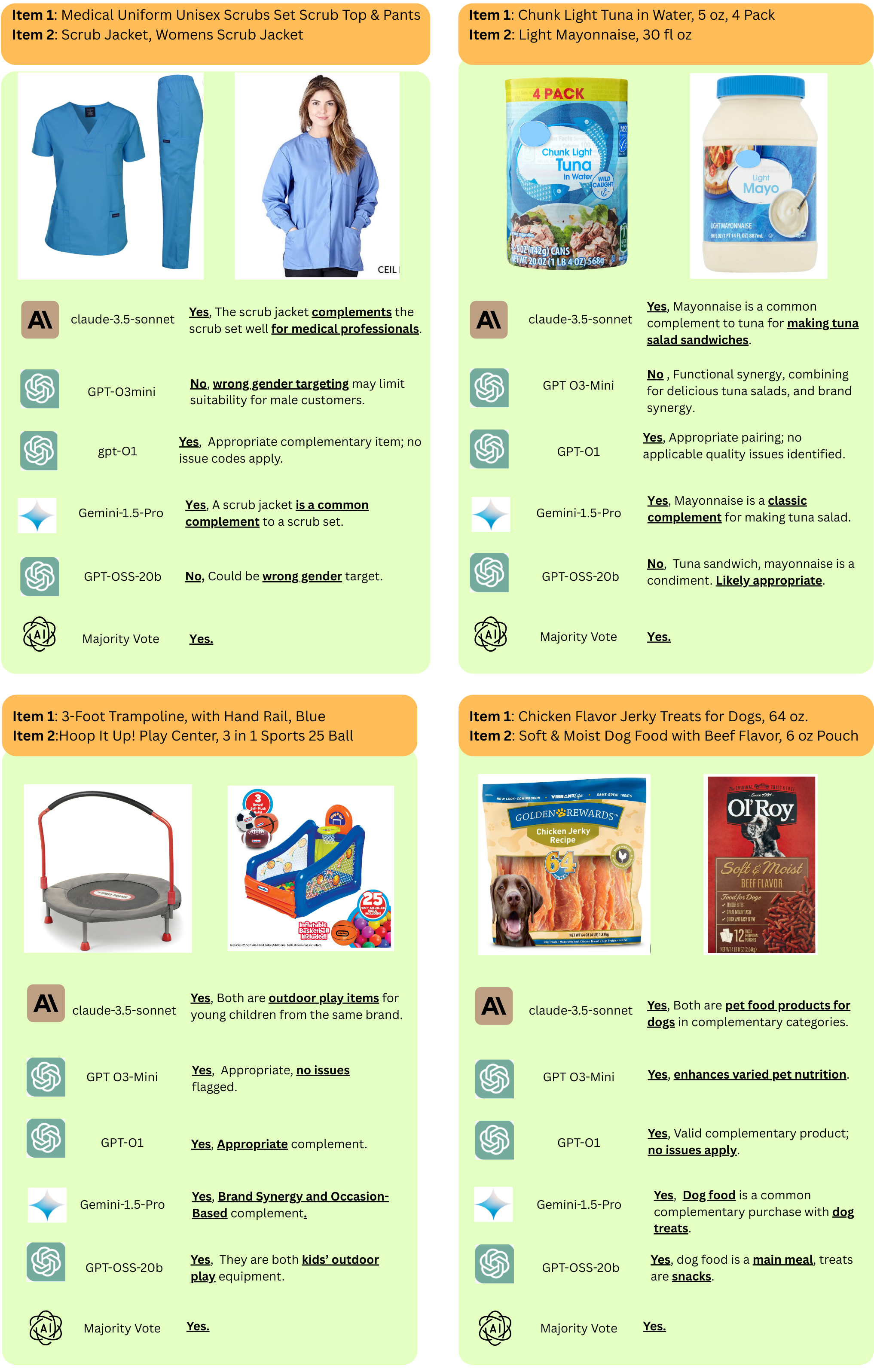} \caption{Example of item pairs from Clothing, Food, Toys, and Pet Supplies and different LLM's output evaluation response for them.}
    \label{fig:case_studies}
\end{figure}

Figure~\ref{fig:case_studies} presents representative anchor–recommendation pairs across four domains. In medical apparel, a scrub set paired with a scrub jacket was consistently judged as complementary by Claude-3.5-sonnet and Gemini-1.5-Pro, emphasizing professional usage, while open-source models flagged possible gender-targeting issues. Similarly, in food and beverages, tuna and mayonnaise were widely recognized as canonical complements for tuna salad preparation, though GPT-OSS-20B expressed uncertainty by treating mayonnaise as a condiment. In both cases, majority voting resolved isolated disagreements, underscoring the robustness of consensus-driven aggregation.

For toys and sports equipment, a trampoline and play center were unanimously judged as valid complements, with rationales ranging from brand synergy to functional overlap in outdoor play. In pet supplies, however, pairing dog food with jerky treats revealed subtle disagreements: some models highlighted complementarity in pet nutrition, while others distinguished between staple meals and snacks. Nevertheless, majority voting converged on a positive outcome. Overall, these case studies highlight a consistent pattern: structured domains with well-defined functional relationships (e.g., food preparation, children’s toys) yield strong agreement, whereas lifestyle or hierarchical categories (e.g., apparel, pet products) introduce more model divergence. ScalingEval effectively mitigates such divergences, producing stable, interpretable judgments without human annotation.

\section{Related Work}\label{sec:litreview}

Large language models (LLMs) have rapidly become \emph{evaluation engines} for open-ended tasks, motivating a growing body of work on ``LLM-as-a-judge'' methods and their reliability. Benchmarks such as MT-Bench and Chatbot Arena established that strong LLM judges can approximate expert and crowd preferences using pairwise comparison protocols, while also surfacing systematic biases such as verbosity and position effects \cite{zheng2023judging,chiang2024chatbot}. Open evaluator models (e.g., Prometheus/Prometheus~2) further pursue transparent, rubric-driven evaluation to improve reproducibility and cost \cite{kim2023prometheus,kim2024prometheus}. At the same time, meta-evaluation studies and surveys caution that judge models may favor their own generations, exhibit family-specific biases, and require debiasing strategies (e.g., length-controlled scoring in AlpacaEval~2) to align with human judgments \cite{dubois2024length,gu2024survey}. 

For recommender systems (RS), these developments are timely: evaluation quality has long hinged on choices of protocol, metric, and data construction \cite{maragheh2025future, maragheh2025arag}. Classic works for recommendation system evaluation distinguish offline from online A/B testing \cite{herlocker2004evaluating,gunawardana2012evaluating}, highlights that accuracy of recommendation alone is insufficient for user value \cite{mcnee2006being}, and documents how evaluation protocol choices and dataset bias can mislead conclusions \cite{canamares2020offline,castells2022offline, ferrari2019we}. These suggest that LLM-based judges can complement (rather than replace) established RS evaluation by providing scalable, semantics-aware assessments when protocols control for known biases.

Majority voting is a long-standing heuristic for synthesizing judgments; foundational models such as \cite{dawid1979maximum,whitehill2009whose} improve on it by jointly inferring instance difficulty and annotator expertise. Theoretical analyses of task assignment and label aggregation further characterize when and how redundancy and weighting outperform simple voting \cite{karger2014budget}. In recommender evaluation, aggregation issues surface both in human studies (e.g., topical or task difficulty effects) and in the construction of ``ground truth'' from behavioral logs, where co-purchase signals can be confounded by substitutes, trends, or merchandising \cite{mcauley2015inferring}. For complementary-item recommendation (CIR) in particular, recent work shows that co-purchase labels are often noisy proxies for functional complementarity; frameworks like P-Companion explicitly separate relevance and diversity, while NEAT proposes label-noise-resistant modeling and trustworthy label generation, and newer analyses revisit behavior-based labels against function-based criteria \cite{hao2020p,ma2021neat,sugahara2024really}. These findings underscore that evaluation pipelines benefit from ``consensus mechanisms'' that are robust to annotator/model idiosyncrasies and to noisy behavioral signals which is precisely the gap that principled label aggregation aims to fill.

ScalingEval positions LLMs themselves as judges and integrates them with Recommender System-aware evaluation practice. Methodologically, we decompose CIR auditing into specialized agents, enforce conflict-resolution rules, and synthesize consensus labels across diverse model families via scalable majority voting inspired by classical aggregation (but applied to models rather than crowd workers) \cite{dawid1979maximum,whitehill2009whose,karger2014budget}. This consensus is then used to benchmark 36 LLMs on large-scale CIR data, yielding category-sensitive insights. Our evaluation framework complements offline recommender system evaluation \cite{schnabel2016recommendations, swaminathan2017off} with an ``LLM-judge consensus'' layer, and complements LLM-judge literature by (i) grounding tasks in a real RS setting (CIR), (ii) reporting cost–latency–accuracy trade-offs across families, and (iii) providing a reproducible, auditor-orchestrated procedure that mitigates known judge biases via multi-model aggregation and category-level analysis.

\clearpage
\section{A Demo of The Evaluation Report Generation}
In this section, we outline the complete process for generating the evaluation report and the details in prompt design.
At the beginning, the user starts a query \textit{"Write an evaluation report for CI recommendation pairs"} and provides a list of recommendations. CI pattern audit prompt and the recommendation issue audit prompt are conducted on each task pair. Finally, all pairs along with the audit decision in previous steps are collected and summarized by the report generation prompt into a well-structured analysis report.

\begin{tcolorbox}[mybox,title={\textit{CI Pattern Audit Prompt}}]
\system\ 

\small{
We propose the following CI patterns to determine if the recommended item is complementary to the anchor item.

1. ACCESSORY OR ADD-ON \\
– Definition: The recommended item augments or protects the anchor.\\ 
– Examples: A phone case or screen protector for a smartphone.\\

2. REPLENISHMENT OR CONSUMABLE \\
– Definition: The recommended item is used up alongside, or as a necessary refill for, the anchor. \\
– Examples: Ink cartridges for a printer.\\

3. FUNCTIONAL SYNERGY \\
– Definition: The recommended item and anchor combine to deliver a fuller or improved functionality. \\
– Examples: Camera lens or tripod for a DSLR camera.\\

4. AESTHETIC OR STYLE MATCH \\
– Definition: The recommended item complements the look or style of the anchor (often fashion- or décor-related). \\
– Examples: A matching scarf for a coat.\\

5. BUNDLED SOLUTION OR “COMPLETE THE SET” \\
– Definition: The recommended item helps complete a set or create a bundled offer. \\
– Examples: A bed frame to go with a matching headboard \\

6. BRAND SYNERGY \\
– Definition: Both the anchor and the recommended item come from the same brand or collection, ensuring consistent quality or compatibility. \\
– Examples: Matching laptop charger or accessory from the same manufacturer \\

7. OCCASION-/USE-CASE-BASED COMPLEMENT \\
– Definition: Items that pair well together for a specific event, activity, or purpose. \\
– Examples: Camping gear (tent + sleeping bag) \\

8. Other
}\\

\user\ 

You are an eCommerece specialist. Your expertise is to evaluate if the recommended item is complementary to the anchor or not. 
If complementary, determine the CI patterns for the given a pair of products. Limit the answer to 15 words.\\
The item pairs are: \\
\textbf{anchor item}: \textit{\{anchor\_title\}}\\
\textbf{recommended item}: \textit{\{recomended\_title\}}
\\

\end{tcolorbox}

\clearpage
\begin{tcolorbox}[mybox,title={\textit{Recommendation Issue Audit Prompt}}]
\system\ 

\small{
You are an e-commerce merchandising specialist responsible for auditing complementary product recommendations.
For each candidate recommendation, decide whether it is appropriate for the anchor product.\\
We have provide typical issues that can be used to evaluate the recommendation items:

1. Accessory / Refill for a Different Product

Definition: Designed for another model, device, or incompatible product.

Example: Recommending iPhone 14 case for an iPhone 12 anchor.\\

2. Embarrassing or Sensitive Content

Definition: May cause discomfort in a public shopping or gift-giving context.

Example: Recommending adult diapers alongside a children’s toy.\\

3. Product Category Too Distant

Definition: Minimal or no functional, thematic, or usage relationship.

Example: Recommending motor oil for a kitchen blender anchor.\\

4. Too Similar to Anchor

Definition: Substitute rather than a complement.

Example: Recommending Pepsi when the anchor is Coca-Cola (competing substitutes).\\

5. Wrong Age / Gender Targeting

Definition: Mismatch in intended demographic.

Example: Recommending toddler shoes for a men’s dress shirt.\\

6. Wrong Format

Definition: Different media/format type that doesn’t complement usage.

Example: Recommending an eBook when the anchor is a physical bookstand.\\

7. Wrong Size or Dimensions

Definition: Physically incompatible measurements or capacity.

Example: Recommending queen-size bed sheets for a twin bed.\\

8. Other

Definition: Any issue not covered above.

Example: Recommending a discontinued or unavailable item.\\

In the task of complementary recommendation, what other issues are not covered? If any,please update the issues list.
}\\

\user\ 

You are a complementary product quality reviewer for an e-commerce catalog. Given an anchor product and 
a candidate recommended product, decide whether the recommendation is appropriate as a complement and 
flag any applicable issue codes. Limit the answer to 15 words.
The item pairs are: \\
\textbf{anchor\_item}: \\
title: \textit{\{anchor\_title\}},\\
product\_type: \textit{\{anchor\_product\_type\}},\\
product\_category: \textit{\{anchor\_product\_category\}}\\
\textbf{recommended\_item}: \\
title: \textit{\{recommended\_title\}},\\
product\_type: \textit{\{recommended\_product\_type\}},\\
product\_category: \textit{\{recommended\_product\_category\}}
\\

\end{tcolorbox}

\clearpage
\begin{tcolorbox}[mybox,title={\textit{Report Generation Prompt}}]
\system\ 

\small{
You are an analytics assistant tasked with generating a report based on the following JSON configuration:

\begin{verbatim}
{
    "action": "generate_report",
    "normalization_rules": {
        "reject_terms": [
            "not complementary",
            "inappropriate",
            "no functional relationship",
            "no CI patterns"
        ],
        "good_terms": [
            "appropriate",
            "complementary",
            "works well",
            "ideal for"
        ],
        "issue_hints": {
            "ACC-DIFF": "Accessory / Refill for Wrong Product",
            "SENS": "Sensitive / Embarrassing Content",
            "CAT-DIST": "Category Too Distant",
            "SUBST": "Too Similar / Substitute",
            "WRONG-AGE-GEN": "Wrong Age / Gender Targeting",
            "WRONG-FORMAT": "Format Mismatch",
            "WRONG-SIZE-DIM": "Size / Dimension Mismatch",
            "OTHER": "Other Issue"
        },
        "conflict_resolution": 
            "If mixed signals, choose the *stricter* judgement 
             (Reject > Major > Minor > Good) and mark 
             pair as conflicted=true."
    },
    "report_requirements": {
        "compute_agreement": true,
        "compute_issue_frequency": true,
        "identify_conflicted_pairs": true,
        "suggest_new_issue_codes": true
    },
    "data_pairs": []
}
\end{verbatim}

\user\ 

Please analyze the data and generate a structured report based on the provided configuration.
\\
}
\end{tcolorbox}


\small
\begin{tcolorbox}[reportbox,title={Recommendation Evaluation Analysis Report Demo}]

This report analyzes a dataset of product recommendations following a defined set of normalization rules and requirements. The analysis includes agreement rates, frequency of issues encountered, identification of conflicted pairs, suggestions for new issue codes, and recommendations for prompt revisions.

\medskip
\hrule
\medskip

\section*{1. Agreement Analysis}
Total Data Pairs Analyzed: 100

Recommendation Audit Results
\begin{center}
\begin{tabular}{|l|c|}
\hline
\textbf{Audit Type} & \textbf{Count} \\
\hline
Appropriate Recommendations   & 64 \\
Inappropriate Recommendations & 18 \\
Too Similar to Anchor         & 7  \\
Product Category Too Distant  & 11 \\
Not Complementary             & 15 \\
\hline
\end{tabular}
\end{center}

Agreement Rate Calculation
\begin{itemize}
  \item \textbf{Agreement Rate} = (Number of Appropriate Recommendations / Total Data Pairs) $\times$ 100
  \item \textbf{Agreement Rate} = (64 / 100) $\times$ 100 = \textbf{64\%}
\end{itemize}

\medskip
\hrule
\medskip

\section*{2. Issue Frequency Analysis}

Summary of Issues Found
\begin{center}
\begin{tabular}{|l|c|}
\hline
\textbf{Issue Type} & \textbf{Frequency} \\
\hline
Appropriate Recommendations  & 64 \\
Inappropriate Recommendations & 18 \\
Too Similar to Anchor        & 7  \\
Product Category Too Distant & 11 \\
Not Complementary            & 15 \\
Complementary Recommendations & 83 \\
\hline
\end{tabular}
\end{center}

Notable Observations
\begin{itemize}
  \item A significant number of appropriate recommendations points to solid performance.  
  \item Presence of inappropriate recommendations and conflicts indicates areas for improvement.
\end{itemize}

\medskip
\hrule
\medskip

\section*{3. Identification of Conflicted Pairs}

Below are the pairs that exhibited conflicts based on the stricter judgment rules:

\begin{center}
\renewcommand{\arraystretch}{1.2}
\begin{tabularx}{\textwidth}{|>{\raggedright\arraybackslash}X|>{\raggedright\arraybackslash}X|>{\raggedright\arraybackslash}X|}
\hline
\textbf{Anchor} & \textbf{Recommendation} & \textbf{Conflict Reason} \\
\hline
NBA 2K20 Legend Edition (PS4) & Red Dead Redemption 2 (PS4) & Too Similar to Anchor \\
\hline
Roku Voice Remote & Roku Smart Home Indoor Smart Plug SE & Product Category Too Distant \\
\hline
Roku Premiere 4K/HDR Player & onn. Android TV 4K UHD Streaming Device & Not Complementary \\
\hline
DuraPro TV Wall Mount & Sceptre 65'' 4K UHD LED TV & Too Similar to Anchor \\
\hline
Samsung Galaxy Watch (42 mm) & Google Nest Hub (Charcoal) & Product Category Too Distant \\
\hline
Apple AirPods (Wireless) & Gourmia 8-Qt Digital Air Fryer & Product Category Too Distant \\
\hline
Beats Studio3 Headphones & Beats Flex Wireless Earphones & Too Similar to Anchor \\
\hline
Apple AirPod Pro (Replacement) & Apple Pencil (1st Gen) & Product Category Too Distant \\
\hline
onn. 64GB SDXC Card & onn. 50'' Tripod & Product Category Too Distant \\
\hline
\end{tabularx}
\end{center}

\medskip
\hrule
\medskip

\section*{4. Suggested New Issue Codes}

\begin{itemize}
  \item \textbf{COMPAT}: Compatibility Issue — items appear complementary but do not function well together.  
  \item \textbf{FUNC-MIS}: Functional Mismatch — items differ in intended application.  
  \item \textbf{CTX-MIS}: Contextual Suitability Issue — items do not match the context, season, or usage.  
\end{itemize}

\medskip
\hrule
\medskip

\section*{Conclusion}

The analysis reports a 64\% agreement rate within the recommendations, reflecting a robust framework for product pair evaluations.  
The presence of a non-negligible proportion of conflicts suggests a need for refined guidelines and the introduction of new issue categories to improve future recommendations.

\textbf{Recommended Next Steps}
\begin{itemize}
  \item Implement the new issue codes for better classification.  
  \item Create detailed annotation guidelines based on findings.  
  \item Address identified conflicts to improve recommendation suitability and customer satisfaction.  
\end{itemize}

\clearpage

\end{tcolorbox}

\end{document}